%% file: main.tex
\documentclass{article}
\usepackage[final]{nips_2016}

\pdfoutput=1

\usepackage{url}

\title{End-to-End Eye Movement Detection Using\\Convolutional Neural Networks}

\author{Sabrina Hoppe\\Perceptual User Interfaces Group\\Max Planck Institute for Informatics, Germany\\\texttt{shoppe@mpi-inf.mpg.de}
\And Andreas Bulling\\Perceptual User Interfaces Group\\Max Planck Institute for Informatics, Germany\\\texttt{bulling@mpi-inf.mpg.de}
}

\newcommand{\cnn}{\textsc{CNN}}
\newcommand{\cnns}{\textsc{CNN}s}

\usepackage{tikz}
\usetikzlibrary{intersections,calc}
\usetikzlibrary{decorations.pathreplacing}

\usepackage{listings}
\usepackage{microtype}
\usepackage{booktabs}

\begin{document}

\maketitle

\begin{abstract}

Common computational methods for automated eye movement detection -- i.e.\ the task of detecting different types of eye movement in a continuous stream of gaze data -- are limited in that they either involve thresholding on hand-crafted signal features, require individual detectors each only detecting a single movement, or require pre-segmented data.
We propose a novel approach for eye movement detection that only involves learning a single detector end-to-end, i.e.\ directly from the continuous gaze data stream and simultaneously for different eye movements without any manual feature crafting or segmentation.
Our method is based on convolutional neural networks (CNN) that recently demonstrated superior performance in a variety of tasks in computer vision, signal processing, and machine learning.
We further introduce a novel multi-participant dataset that contains scripted and free-viewing sequences of ground-truth annotated saccades, fixations, and smooth pursuits.
We show that our CNN-based method outperforms state-of-the-art baselines by a large margin on this challenging dataset, thereby underlining the significant potential of this approach for holistic, robust, and accurate eye movement protocol analysis.

\end{abstract}

\input{intro}

\input{relatedwork}

\input{method}

\input{dataset}

\input{results}

\input{conclusion}

\section*{Acknowledgements}

This work was funded, in part, by the Cluster of Excellence on Multimodal Computing and Interaction (MMCI) at Saarland University as well as a PhD scholarship by the German National Academic Foundation.

\bibliographystyle{apalike}
\bibliography{refs}
\end{document}

%% file: intro.tex
\section{Introduction}

Eye movement detection is fundamental for automated visual behaviour analysis and important for many gaze applications, most importantly in human-computer interaction and experimental psychology~\cite{salvucci2001automated}.
For example, the movements that our eyes perform throughout the day reflect our activities and can be used for eye-based activity recognition~\cite{bulling11_pami}.
In addition, eye movements are closely linked to visual information processing and cognition.
Consequently, previous works demonstrated eye-based prediction of cognitive processes, such as visual memory recall or cognitive load~\cite{stuyven2000effect,bulling11_ubicomp}, fatigue~\cite{schleicher2008blinks}, as well as personality traits~\cite{risko2012curious,hoppe15_ubicomp} that are difficult if not impossible to assess using other modalities.

Despite significant advances in high-level visual behaviour modelling and analysis, developing a single detector for low-level eye movements remains challenging~\cite{komogortsev2013automated}.
This is, in part, due to the non-stationary nature and thus unpredictability of visual behaviour, but also because of noise that is superimposed on the gaze data as well as significant differences and person-specific variability in eye movement characteristics.

\begin{figure}
    \centering
    \includegraphics[width=0.8\textwidth]{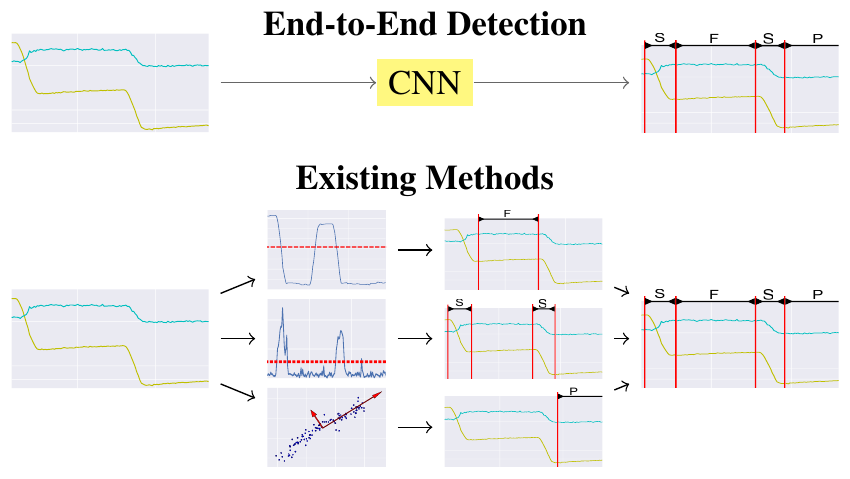}
    \caption{Our method based on convolutional neural networks (CNN) learns a single eye movement detector directly from the continuous two-dimensional gaze data stream. In contrast, existing methods require multiple processing stages to classify individual eye movements.}
    \label{fig:teaser}
\end{figure}

A large body of work focused on methods for detecting fixations or saccades~\cite{salvucci2000identifying}.
In contrast, detection of smooth pursuits or vestibulo-ocular reflexes remains relatively unexplored~\cite{vidal11_petmei,vidal2012_etra,larsson2014discrimination}.
However, existing methods for eye movement detection are limited in that they 1) often require thresholding on manually defined signal features, 2) only detect a single eye movement type, and 3) only perform classification, i.e.\ assume pre-segmented gaze data that is then classified into the desired target eye movement.
Another limitation for the development of methods for eye movement detection is the lack of publicly available algorithm implementations and eye movement datasets.

The goal of this work is to address both of these limitations.
To this end we propose a fundamentally different approach for eye movement detection that involves learning a single detector end-to-end, i.e.\ directly from raw gaze data to the different types of eye movement.
Our method is based on convolutional neural networks (CNN), a specific type of neural network that recently demonstrated superior performance for diverse tasks such as image classification, appearance-based gaze estimation, or human activity recognition~\cite{imagenet,zhang15_cvpr,yangdeep}.
In contrast to existing methods, our end-to-end method directly maps from a continuous stream of input gaze data to eye movement labels for all three classes without segmentation steps and or any human intervention. 
In addition to the eye movement type, the method also returns a confidence value for each gaze sample that can be used for probabilistic modelling and analysis of human visual behaviour.
To facilitate algorithm developments by the research community and enable principled comparisons of different algorithms, we further introduce a novel multi-participant dataset that contains scripted and free-viewing sequences of ground-truth annotated saccades, fixations, and smooth pursuits.
The full dataset including annotations can be made available upon request.

The specific contributions of this work are threefold.
First, we present the first method to detect three key eye movements, namely fixations, saccades and smooth pursuits, simultaneously from a continuous sequence of gaze samples as returned by current eye trackers.
The method is based on a convolutional neural network that does not require thresholding, data pre-segmentation, or any other manual intervention.
Second, we introduce a novel multi-participant dataset that contains a diverse set of scripted and free-viewing sequences of saccades, fixations, and smooth pursuits.
In total, the dataset includes 1,626 fixations, 2,647 saccades and 1,089 pursuit movements that were manually annotated  and correspond to about 400,000 frames of high-frequency gaze data.
Third, we demonstrate that our method outperforms state-of-the-art baselines by a large margin on this challenging dataset, thereby underlining the significant potential of this new approach for robust and accurate eye movement detection.

%% file: relatedwork.tex
\section{Related Work}

The first types of eye movements to be investigated were saccades and fixations which occur when people are viewing static scenes, for instance images or text (cf. for instance ~\cite{mcconkie1981evaluating}).
Many more algorithms were to follow until today and are typically seen in three categories (e.g. in \cite{salvucci2000identifying,holmqvist2011eye,kasneci2015online}): velocity-based methods, dispersion-based methods and probabilistic methods. 

\subsection*{Velocity-Based Methods}

Saccades are fast eye movements while fixations are rather stable and thus characterised by a low velocity. 
Velocity-based methods thus classify everything above a fixed velocity threshold as saccades and everything below as fixations~\cite{salvucci2000identifying}. 
Since smooth pursuits are typically faster than fixations, but not as fast as saccades, this idea can be extended to the three class problem by introducing a second threshold~\cite{ferrera2000task,komogortsev2013automated}.
However, velocities of eye movements vary between people and even withing people for different tasks, which makes it hard to find generally well performing thresholds and renders the algorithms hardly usable for real-life scenarios with dynamic scenes ~\cite{kasneci2015online}.
If smooth pursuit is considered as a third type of eye movement, the problem becomes even harder as pursuits are known to be affected more severely by factors such as alcohol or fatigue ~\cite{bahill1980smooth}.

\subsection*{Dispersion-Based Methods}

Dispersion-based algorithms exploit that points belonging to fixations cluster around one position, while points belonging to fixations or pursuits are spread out farther ~\cite{holmqvist2011eye}. 
Different formulations of this intuition have been proposed, e.g. the difference between the smallest and largest x or y component of a gaze signal ~\cite{blignaut2009fixation} or dispersion-related properties of the minimum spanning tree between all gaze points within a time window ~\cite{salvucci2000identifying}.

There are also proposals to epxloit dispersion properties of the data indirectly: Either by using clustering methods based on the spatial position of points, e.g. using projection clustering ~\cite{urruty2007detecting} or using Principle Component Analysis which finds those dimensions along which a set of data points varies most ~\cite{kasneci2015online}.
Combining similar eigenvector analyses with velocity scores has been proposed by \cite{berg2009free}.
Although both velocity- and dispersion based methods seem to work for many specific applications, it has been shown how the usage of velocity and dispersion thresholds can systematically bias the result of eye tracking studies \cite{shic2008incomplete}.
Further detailed geometric properties of the trajectory such as slopes and integrals of the signal components~\cite{vidal2012_etra} can be exploited to classify fixations and smooth pursuits -- at least given pre-segmented data.

Most of the more simplistic methods suffer from similar issues as velocity-based approaches, namely the difficulty to find generally valid thresholds.
PCA-based methods seem to generalise and perform better, although they also use an empirically set threshold to distinguish between principle components for fixations from those for pursuits.
Additionally, all of the named algorithms in this section either only work for two classes or need to tackle the three-class problem in two stages, each of which solves a binary task.
Our approach can overcome both limitations since it does not require segmented data, no fixed threshold needs to be set and gaze data can be labelled as one of three classes in a single prediction step.

\subsection*{Probabilistic Methods}
 
As another way to overcome these problems, probabilistic methods have been proposed and were found to distinguish better between fixations and saccades ~\cite{salvucci2000identifying,kasneci2014applicability}. 
For instance, Hidden Markov Models (HMM) can successfully classify eye movements as fixations or saccades ~\cite{tafaj2012bayesian}.
They have also been shown to be particularly suitable for on-line analyses as HMMs can handle successively arising data points naturally ~\cite{kasneci2015online}.
Another probabilistic method proposed to distinguish between fixations and saccades are Kalman filters ~\cite{sauter1991analysis} which have been applied successfully to gaze user interfaces ~\cite{komogortsev2007kalman}.
However, none of these probabilistic models can distinguish between all three types of gaze data in a single step.

\subsection*{Other Methods}

One way of simplifying the problem is of course to take additional information into account. For instance, if a set of potentially followed targets is known, it is easier to detect smooth pursuits by matching gaze to stimulus trajectories ~\cite{vidal2013pursuitsb}.
A sequence of diverse hand-crafted steps, including outlier detection, pole modelling, segmentation based on shape features as well as a frequentist statistical test, has been proposed as a pipeline to first detect fixations, saccades and postsaccadic oscillations and then intervals previously classified as fixation are reconsidered for pursuit detection ~\cite{larsson2014discrimination,larsson2015detection}. 
Although this constitutes one single pipeline, its outcome on a three-class problem has never been evaluated overall.
Also in contrast to our proposed method, this pipeline is only applicable in off-line analyses as the whole dataset needs to be iterated several times.

%% file: method.tex
\section{End-to-End Eye Movement Detection}

To address the aforementioned limitations we propose to cast the problem of eye movement detection into the problem of  learning a sequence-to-sequence mapping between gaze samples and eye movement labels.
We introduce a method based on deep convolutional neural networks that, from a sequence of gaze samples, predicts a sequence of probabilities for each sample to belong to a fixation, saccade, or smooth pursuit (see Figure~\ref{fig:teaser} for an overview).
Given that this method directly learns a single network instead of performing multiple intermediate processing steps, this kind of learning is also called \textit{end-to-end} learning.
We implemented our model using the Deep Learning library \textsc{Keras}~\cite{keras}.

Different types of eye movement are characterised by the specific ``signal shape'' created by subsequent gaze samples belonging to that movement.
For example, saccades -- the simultaneous fast movements of both eyes that we perform to redirect gaze onto a new target --
can last between 10 and 100 ms~\cite{holmqvist2011eye} and are characterised by a distinct and fast change in amplitude in the gaze data.
In contrast, fixations -- the stationary states of the eyes during which gaze is held upon a specific location in the visual scene -- can, in principle, last for arbitrary amounts of time.
In terms of signal characteristics, they can easily be confused with smooth pursuits -- the movements that the eyes perform when we latch onto a moving object.
While fixations are mostly static, smooth pursuits -- as the name suggests -- are characterised by a continuous, smooth change in the signal over time that needs to be modelled.

\subsection*{Network Architecture}

\begin{figure}
    \centering
    \includegraphics[width=0.7\textwidth]{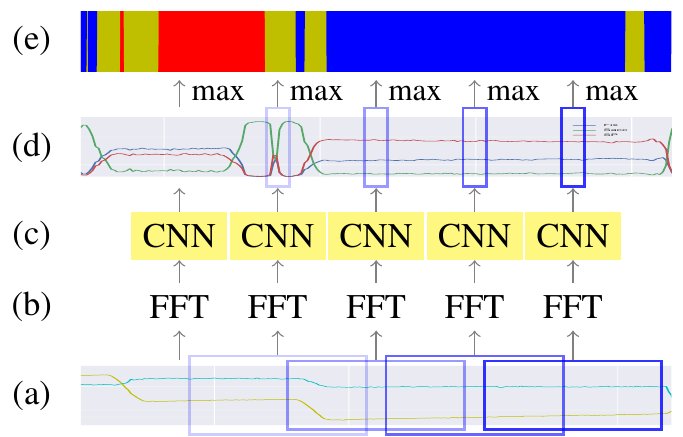}
    \caption{CNN-based eye movement detection: a window from an input stream of raw gaze data (a) is transformed into frequency domain using the fast fourier transform (b). The encoded input is processed by the CNN (c) which calculates probabilities for each eye movement type (d). The movement with maximum probability is predicted (e).}
    \label{fig:pipeline}
\end{figure}

Figure~\ref{fig:pipeline} provides an overview of the proposed method.
Our method takes a continuous stream of two-dimensional gaze samples as input.
To capture relevant eye movement characteristics and accommodate for different eye movement lengths, the stream of gaze samples is always analysed in windows of 30 samples.
To obtain a prediction for each gaze sample, the window moves over the sequence one by one.
We first decompose the signal into different frequency components using a Fast Fourier Transform (FFT).
In this representation, a smooth pursuit shows as a strong low frequency band because the signal changes slowly but steadily over time. 
Gaze jitter during a fixation, however, typically results in a mix of very high frequencies.
The frequency representation of the raw signal is passed on to the CNN which in turn outputs a three-dimensional activation signal. 
Each component of this signal can be interpreted as the probability for one of the three eye movement labels (fixations, saccade, smooth pursuit).
Finally, the label with highest probability can be assigned to the central sample in the window.

\begin{figure}
    \centering
    \includegraphics[width=0.7\textwidth]{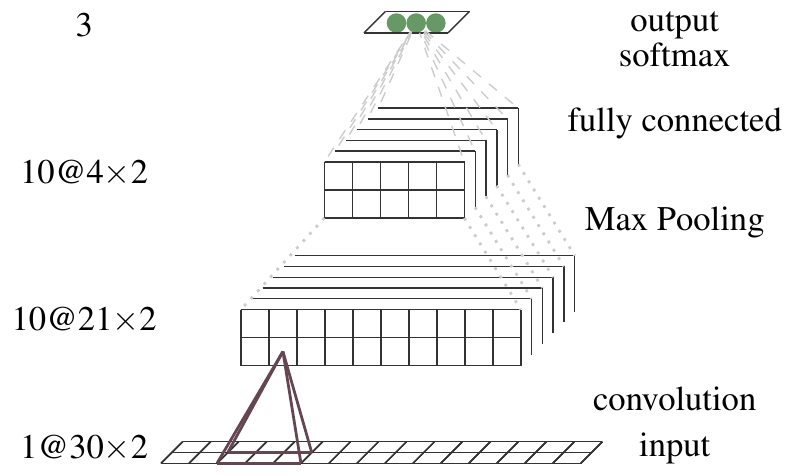}
    \caption{Detailed architecture of the proposed CNN: The input layer is connected to a convolutional layer with 10 filters of kernel size 10$\times$1. On top of that, a max pooling layer is applied with down scale factors 5$\times$1. The output of this layer is fully connected to the three output nodes at which a softmax activation is applied to achieve values that can be interpreted as probabilities. The left column indicates the matrix size for each step.}
    \label{fig:cnn-details}
\end{figure}

The CNN learns to directly predict the three-dimensional output from the frequency representation.
A CNN can be composed of different layers, specifically convolutional layers, max pooling layers, as well as fully connected layers \cite{lecun1998gradient}. 
We evaluated different architectures on a pilot dataset and found the architecture shown in Figure~\ref{fig:cnn-details} to perform best for eye movement detection.
Input to the network is a sequence of gaze samples of size $30\times2$, i.e.\ 30 subsequent gaze samples transformed to a frequency representation, one for horizontal and vertical direction each.
The first layer of the network is a so-called convolutional layer that learns a set of 10 filters. 
Each filter applies its characteristic function to each patch of size $5\times1$ in the input, thereby mapping a neighbourhood of input samples to one value. 
The function used for this convolution is learned automatically during the training of the network.
On top of the convolution layer, we put a so-called \textit{Max Pooling} layer which essentially shrinks the data:
each cell within the pooling layer represents a whole patch from the layer underneath by a single value which is chosen as the maximum inside its patch.
This compressed representation is mapped to a three dimensional output. 
To do this, a fully connected layer is used, i.e.\ a layer in which each node is connected to all nodes of the layer underneath. 
Each node computes a different function taking all input node activations as arguments. 
Finally, a softmax operation in the output layer scales the activations for one sample to 1 and hence can be interpreted as probabilities of each label. 

\subsection*{Training Procedure}

The network was trained on 75\% of the data set using the Keras implementation of the \textsc{Adam} optimiser \cite{kingma2014adam} which is an efficient algorithm for stochastic optimisation with four intuitive parameters: $\alpha$ is the step size of an optimisation iteration, $\beta_1$ and $\beta_2$ control how much \textsc{Adam} considers past experiences in the current time step.
More precisely, they are exponential decay rates for the moving average over the first two moments of the gradient along which we want to optimise. 
Additionally, there is also a parameter $\epsilon$ that determines how often solutions that seem non-optimal at first glance are tried again.

First, the network was trained for 100 iterations with $\alpha=0.001$, $\beta_1=0.9$, $\beta_2=0.99$, and $\epsilon=e1-8$. 
After each iteration, the resulting network weights were tested on the validation set, 12.5\% of the data. 
The network with the best performance on the validation set was trained on the same data for another 200 iterations, but this time with parameters  $\alpha=0.002$, $\beta_1=0.85$, $\beta_2=0.1$ and $\epsilon=e1-8$.

All parameters were determined empirically alongside with the model selection process: 
We varied the \cnn{} architecture (i.e. for instance number of filters, number of layers) as well as the window size of the input and finally chose the network performing best on our validation set.
This network was then tested again on the test set, another 12.5\% of the data. 
These results are the ones reported here.

\subsection*{Sample Output}

\begin{figure}[t]
\centering
\includegraphics[width=0.9\linewidth]{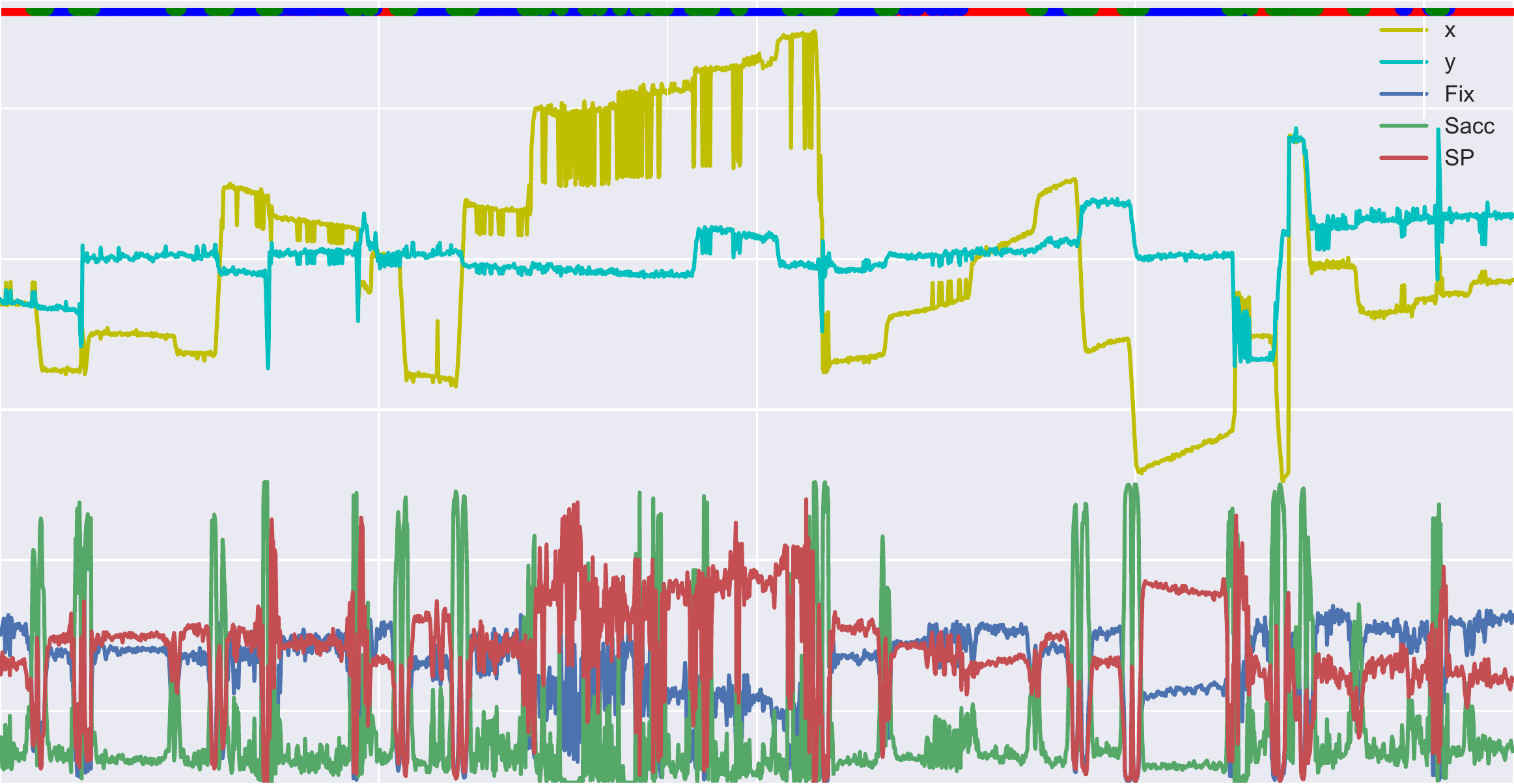}
\caption{Given the gaze stream on top, the activation is plotted on the lower half of the figure. Each colour of the activation stand for the probability one class of eye movements - i.e. at each time frame, the prediction would be the class corresponding to the highest curve. The colours in the top line indicate ground truth labels. }
\label{fig:qualitative}
\end{figure}

A sample activation of the top layer of the CNN together with the original gaze data is shown in Figure~\ref{fig:qualitative}.
In this particular example the green curve, which represents the probability of a saccade, has generally a rather low activation but robustly spikes whenever there is a sudden amplitude change in the gaze data.
The activation for fixations and saccades is close together for many time steps, illustrating that their difference can be very subtle. 
The figure also illustrates cases to distinguish saccades and pursuits for which the x component of the signal (yellow) steadily and noticeably rises.

%% file: dataset.tex
\section{Data Collection}

As we are not aware of any publicly available, fully annotated data set including saccades, fixations and pursuits for a range of tasks, we designed a controlled laboratory study to collect eye movement data in a principled way.
The data collection involved participants in looking at a scripted but randomised sequence of visual dot stimuli triggering diverse saccades, fixations, and smooth pursuits.
We also included variants and combinations of these movements, such as latching onto an already moving dot vs. performing a smooth pursuit from a fixation, or jumping from one moving dot to another moving dot to include saccades between pursuits as well.
Scripting the data collection allowed us to semi-automatically annotate the recorded gaze data post-hoc and therefore to record a much larger number and variations of eye movements.
We finally also included reading as well as images and movies as free viewing stimuli, which we annotated manually.

\def\arraystretch{1.3}%
\begin{table}[th]
\centering
\begin{tabular}{c|cccc}
\textbf{Participant} & \textbf{gender} & \textbf{eye colour} & \textbf{glasses} & \textbf{eye tracking} \\
 			&		&			&			& \textbf{experience}\\
\toprule
1 & male & blue & - & - \\
2 & male & blue & yes & yes\\
3 & male & grey & yes & yes\\
4 & male & brown & yes & yes\\
5 & male & brown & yes & yes\\
6 & male & brown & yes & - \\
7 &female & green &  yes & -\\ 
8 & male & blue &  - & - \\
9 & male & grey & - & yes\\
10 & female & brown & yes & -\\
11 & male & black & yes & yes\\
12 & male & brown & - & yes \\
13 & female & black &yes& -\\
14 & male & black & yes& -\\
15 & female & green & - & -\\
16 & female & brown & yes & - \\
\bottomrule
\end{tabular}
\caption{Overview of the participants in the study.}
\label{table:dataset}
\end{table}

\subsection*{Apparatus}

We used a state-of-the-art Tobii TX300 remote eye tracker that records binocular gaze data at 300 Hz. 
To obtain a single point of gaze on the screen we averaged the estimated gaze points for both eyes.
All participants were sitting comfortably at a fixed distance of about 65cm away from the screen.
No chin rest was used.
The stimulus software was written in \textsc{PsychoPy} \cite{peirce2008generating} and synchronised with the eye tracker using \textsc{PsychoPy}s io hub functionality. 
This allowed for synchronous, timestamped logging of both gaze data from the TobiiTX300 and stimulus information.

\subsection*{Participants and Procedure}

We recruited 16 participants (5 female) between 22 and 32 years through university mailing lists and personal invitations.
11 of our participants normally wear glasses (but not during the recording), 7 of them had active experience with eye tracking in the sense of writing code or analysing data.
Details on our participants are also summarised in Table \ref{table:dataset}.
All participants were paid 10 EUR for participating in the 40 minute data collection.

Upon arrival in the lab participants were first introduced to the purpose of the study, explained the general procedure and visual stimuli, and asked to complete a questionnaire on demographics.
All participants were instructed to move their head as little as possible during the recording but to otherwise behave and, most importantly, follow the visual stimuli as naturally as possible. 
Participants were then calibrated to the eye tracker using a standard 9-point calibration provided by the Tobii Studio software.
Afterwards, the assistant started the experiment software that automatically guided participants through the sequence of visual stimuli.
The experiment was paused for two minutes in the middle to reduce fatigue.

\subsection*{Stimuli}

\begin{figure}
\centering
\begin{minipage}{0.4\linewidth}
\includegraphics[width=\linewidth]{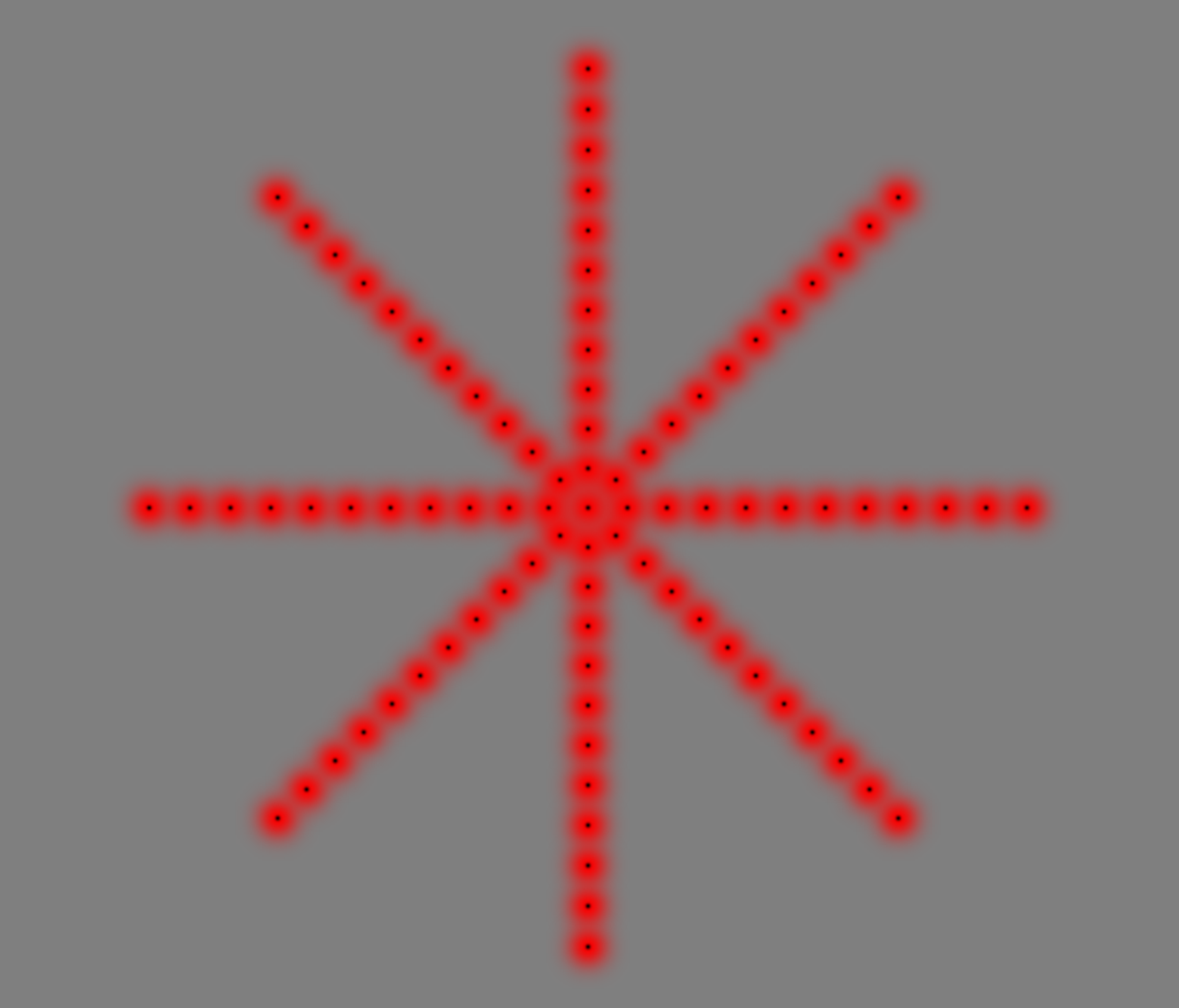}
\end{minipage}
\begin{minipage}{0.4\linewidth}
\includegraphics[width=\linewidth]{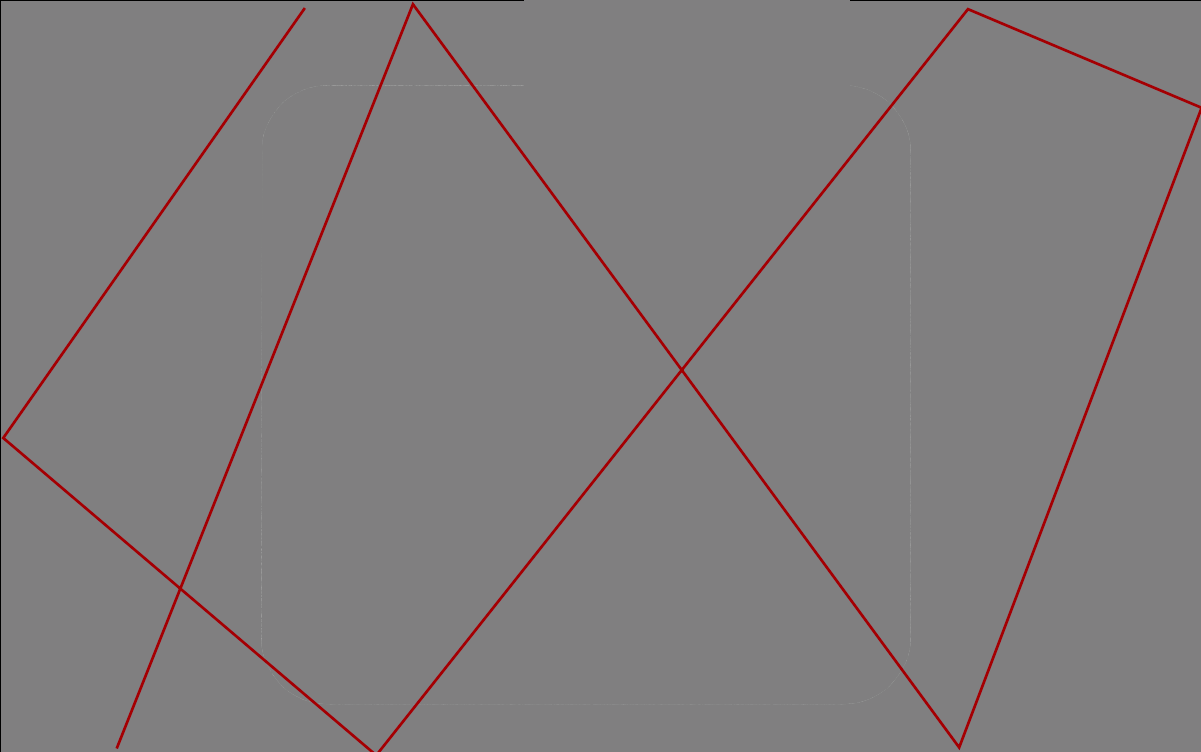}
\includegraphics[width=\linewidth]{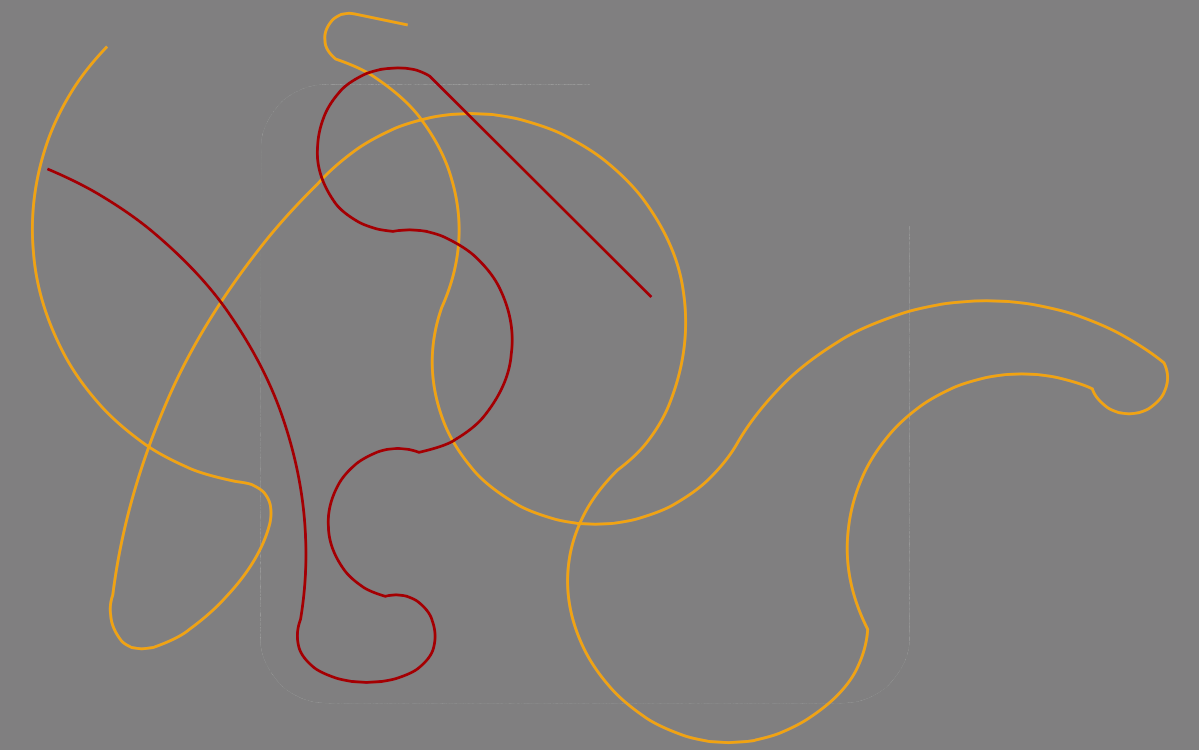}
\end{minipage}
\caption{Start and end positions for saccades and fixations and some of the pursuits were scripted to result in a star-like shape (left). Further pursuits followed random curves, either straight and bouncing back at the edges of the screen or entirely random (right).}
\label{fig:stimuli}
\end{figure}

The experiment contained visual stimuli, images, videos and text to read, which were shown in random order.
The visual stimulus was a small red dot with a radius of 1 degree of visual angle (when located at the center of the screen) moving in front of a light grey background.
The dot stimulus moved according to a scripted sequence to trigger diverse saccades, fixations, and smooth pursuits (see Figure~\ref{fig:stimuli}).
First, a total of 88 positions was computed on a star-like shape resulting from a horizontal line, a vertical line and the two angle bisectors.
The longest distance between two points was 22 visual degrees.
Starting from each of the positions with a fixation of random duration between 100\,ms and 400\,ms, the dot jumped to the point mirrored at the screen center to induce saccadic behaviour.
Between the same pairs of points, slow movements with speeds between 5 and 35 $\deg$/sec were shown to elicit smooth pursuits.
Additionally, five straight movements in random directions were shown for around 10 seconds, while the dot was bouncing back naturally at the edges of the screen.
In some of the conditions, the dot furthermore sped up or slowed down to any speed between 5 and 35 $\deg$/sec.
For further random effects, smooth movements along entirely random trajectories were shown as well as a crowd of up to four points moving randomly, while the only red dot in the scene was to follow.
Colour changes in this set-up created additional saccadic movements between smoothly moving stimuli.
Also there was a set of 10 images, 7 videos and 4 texts that were shown to the user for 5, 60 and 30 to 90 seconds respectively. 
Saccades that were performed to jump between stimuli, for instance from the last fixation on an image to the red dot showing up, were included in the dataset as well.
The stimuli were designed in such a way that including these transitions between stimuli all transitions from any and to any of the three gaze movements are included in the dataset.

\subsubsection*{Annotation}

Data for all stimuli was annotated manually after pilot studies revealed that an automatic annotation based on the stimulus movement leads to many false annotations which seemed to prevent the network from learning.
The annotation was done based on the gaze data, but the type of stimulus was known. 
For instance, while the user was shown static images, no smooth pusuit should have been annotated.
For each of the participants, one random image and one movie were annotated as well as random amounts of data from artificial stimuli and two reading tasks.
It is important to note that the approach proposed here does not contain any hand-crafted features.
It therefore is unlikely that the manual annotation introduced a bias that overestimates performance of our method.

In total, 1,626 fixations, 2,647 saccades and 1,089 pursuit movements were annotated which corresponds to about 400,000 frames of annotated high-frequency gaze data. For training we used 75\% of the data, 12.5\% for the evaluation of different network architectures (the validation set) and the remaining 12.5\% frames were used for the final evaluation (the test set).

%% file: results.tex
\section{Results}

We performed a series of evaluations to better understand the performance of our method and compare it to state-of-the-art approaches.
Given that there is no implementation of the algorithms described in \cite{larsson2014discrimination,larsson2015detection,kasneci2015online} publicly available, we reimplemented the following methods.

\paragraph{Simple dispersion and velocity thresholding.}
Simple dispersion and velocity-based thresholding approaches decide for or against a class depending on whether or not the dispersion/velocity in a given window of gaze data exceeds a defined threshold~\cite{salvucci2000identifying}. 
The corresponding class label is then assigned to all gaze samples inside the window.
The extension of these two algorithms to three classes is a two-step algorithm that first filters saccades based on velocity and then distinguish between fixations and smooth pursuit based on dispersion \cite{komogortsev2010qualitative}.

The Velocity and Movement Pattern Identification (IVMP) algorithm also applies a velocity threshold to filter out saccades. 
As a second step, it uses the mean angle between successive points within a time window to classify between fixations and pursuits ~\cite{komogortsev2013automated}.

\paragraph{PCA-based dispersion thresholding.}
A Principle Component Analysis (PCA) finds dimensions along which data has the most variance. 
Thus, if gaze points in two dimensions are considered, the variance along these two axes becomes informative about the shape of the points:
If they are scattered in a circular shape, it is most likely a fixation, whereas points belonging to a pursuit are rather stretched in one dimension. 
Therefore the ratio of variances in the two PCA dimensions has been proposed as a feature to distinguish between fixations and pursuits \cite{kasneci2015online,larsson2015detection}.

\subsection*{Binary Eye Movement Classification}

Prior work only considered classification, i.e.\ the case for which presegmented gaze data was classified into target or non-target eye movements.
To be able to compare with our detection method, we first studied three binary classification problems.
To create this evaluation setting, we selected one eye movement type as the positive class and all other types as the negative class.
To obtain a complete set of performance values, we iterated over all three eye movement types.
In addition to the classification performance, we also evaluated the relation between prediction accuracy and the probability values associated with the prediction.

\begin{figure*}
\centering
\includegraphics[width=0.31\textwidth]{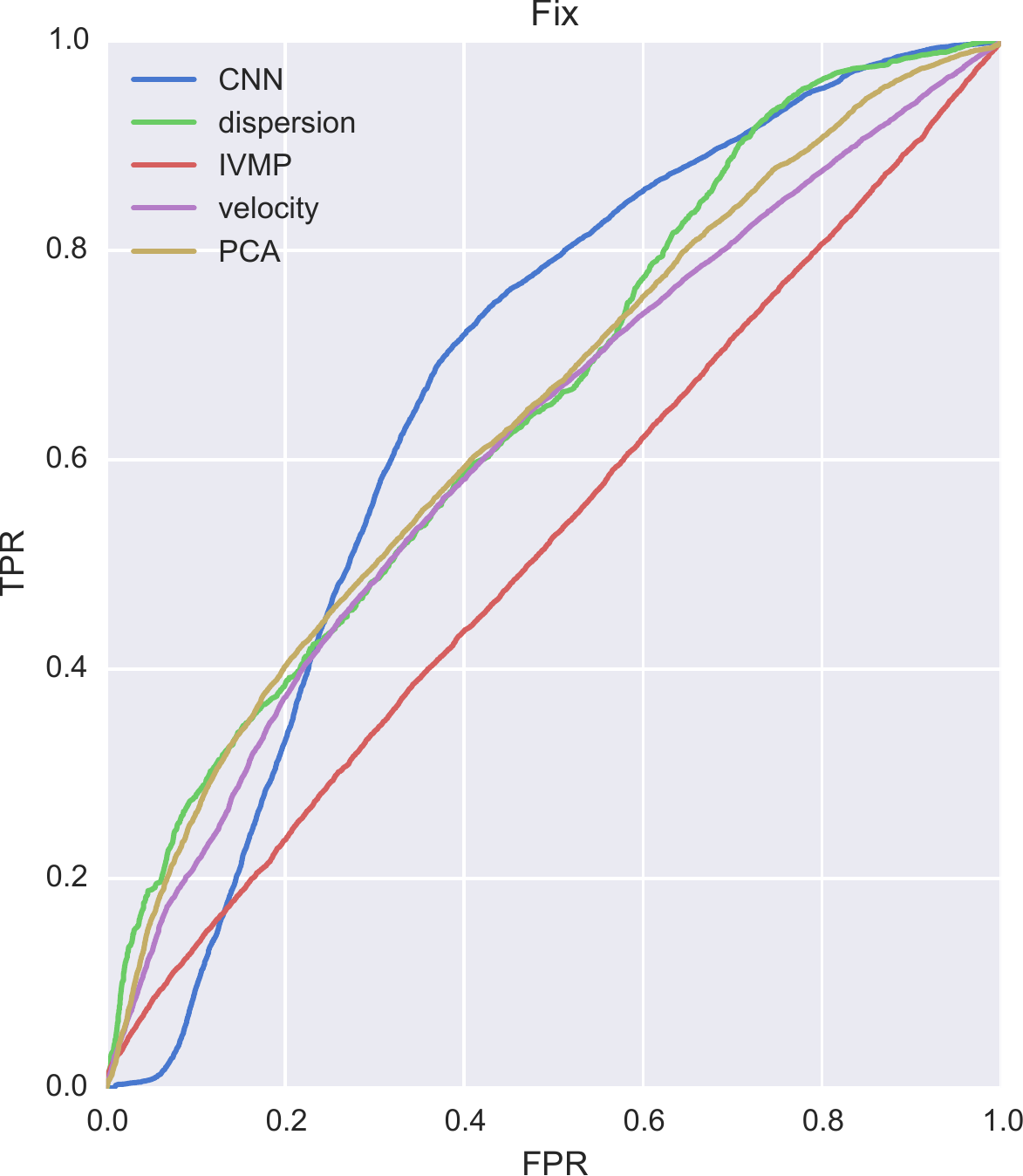}
\hspace{0.2cm}
\includegraphics[width=0.31\textwidth]{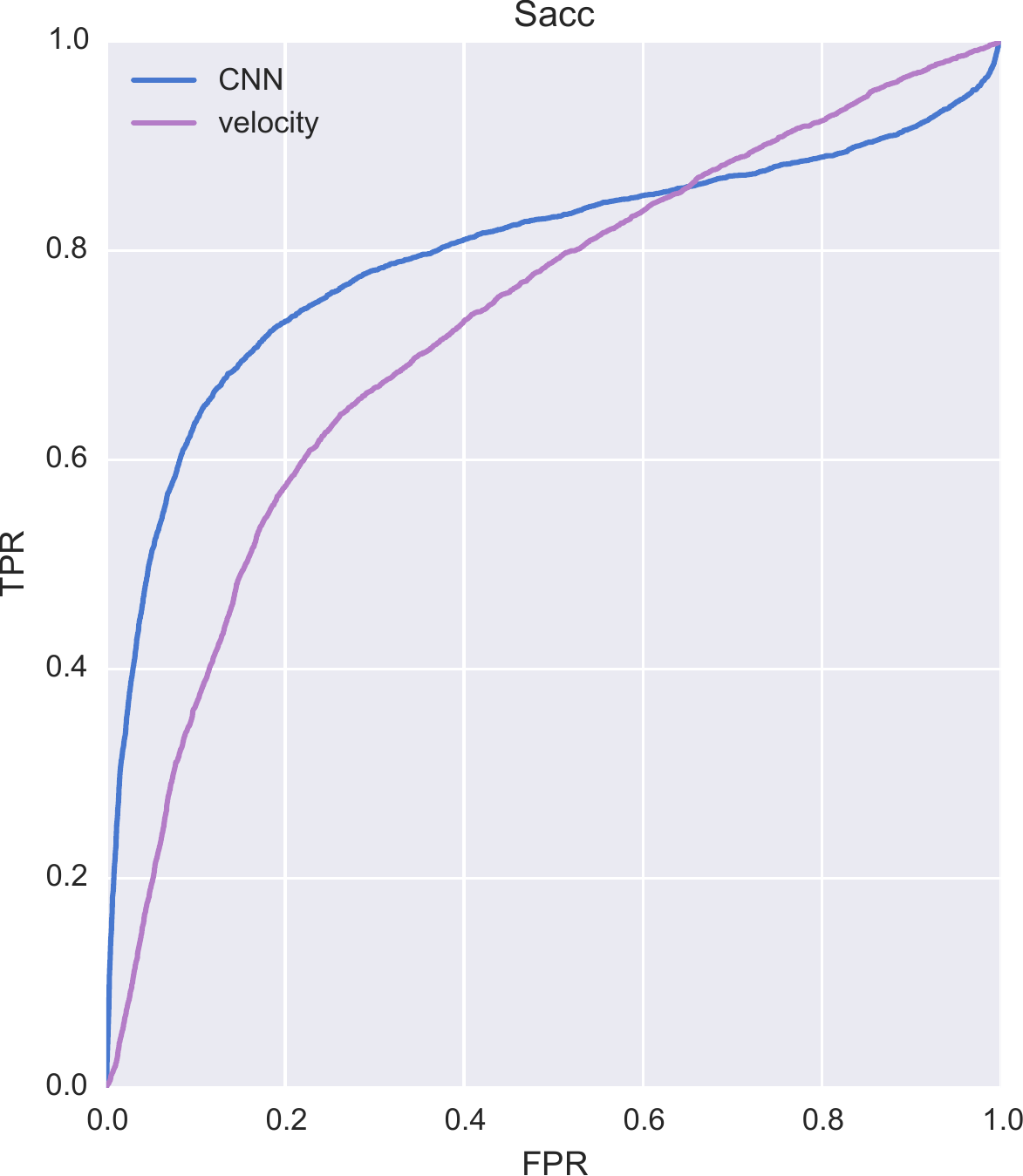}
\hspace{0.2cm}
\includegraphics[width=0.31\textwidth]{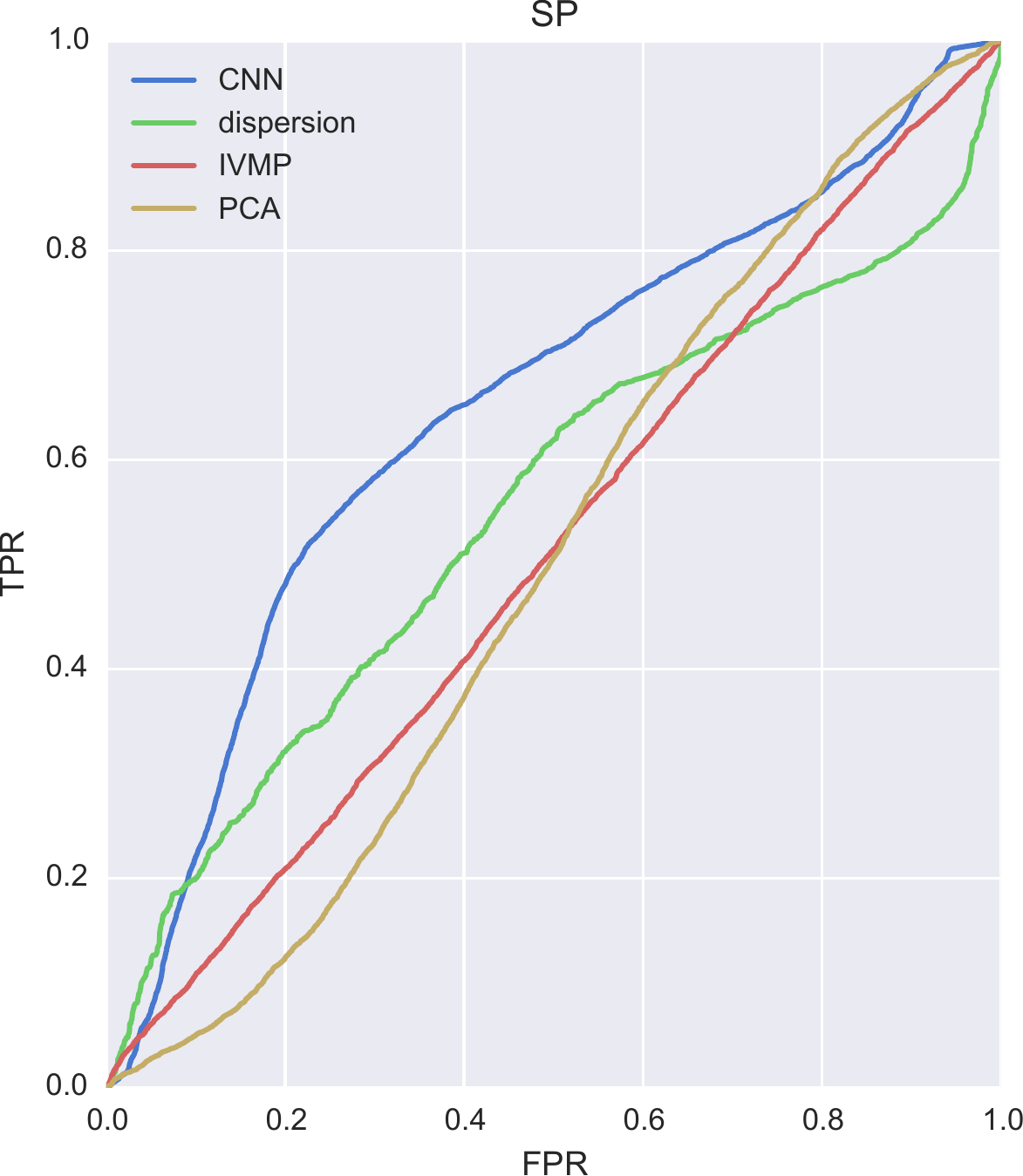}
\caption{Frame-wise performance for separate single class problems: our \cnn{} as well as baseline approaches are evaluated frame by frame. The resulting receiver-operator curves (ROC) are plotted for each of the classes. The \cnn{} achieves areas under the curve (AUC) of 0.67, 0.80 and 0.65 respectively and a mean across classes of 0.71.}
\label{fig:roc}
\end{figure*}

Figure \ref{fig:roc} shows the ROC curves for each of these evaluations. 
As can be seen from the figure, our \cnn{} detection approach outperforms all other methods although it was not trained for this binary classification problem explicitly.
Overall, the method achieves area under the curve (AUC) values of 0.67, 0.80 and 0.65 for fixations, saccades, and pursuits respectively (0.71 AUC on average).

For fixation detection, all other methods could be evaluated and achieve AUCs between 0.53 and 0.65.
For saccades, only the velocity-based thresholding baseline could be evaluated since it is designed for this 1 vs. all classification setting. 
Still, the \cnn{} outperforms the thresholding algorithm on our dataset with the thresholding algorithm achieving an AUC of 0.72, while the \cnn{} achieves 0.80.
The velocity-based approach can not be used for pursuit detection as it first classifies saccades and fixations and only labels the remaining points as pursuits.
However, the dispersion-based algorithm, the IVMP and PCA-based approaches were evaluated and achieved AUCs between 0.51 and 0.55. 
Again our \cnn{} performs substantially better by achieving an AUC of 0.65.

\subsection*{Multi-Class Eye Movement Detection}

The binary classification problems were evaluated to be able to compare to existing approaches.
However, our algorithm is capable of detecting eye movements in the continuous gaze stream and of assigning labels for all three eye movement types simultaneously.
We therefore further evaluated our algorithm separately for the three-class detection problem, first by evaluating predictions sample-by-sample and then event-wise.
Unfortunately, no comparisons with previous methods are possible in this case.

\begin{figure}
\centering
\includegraphics[width=0.45\linewidth]{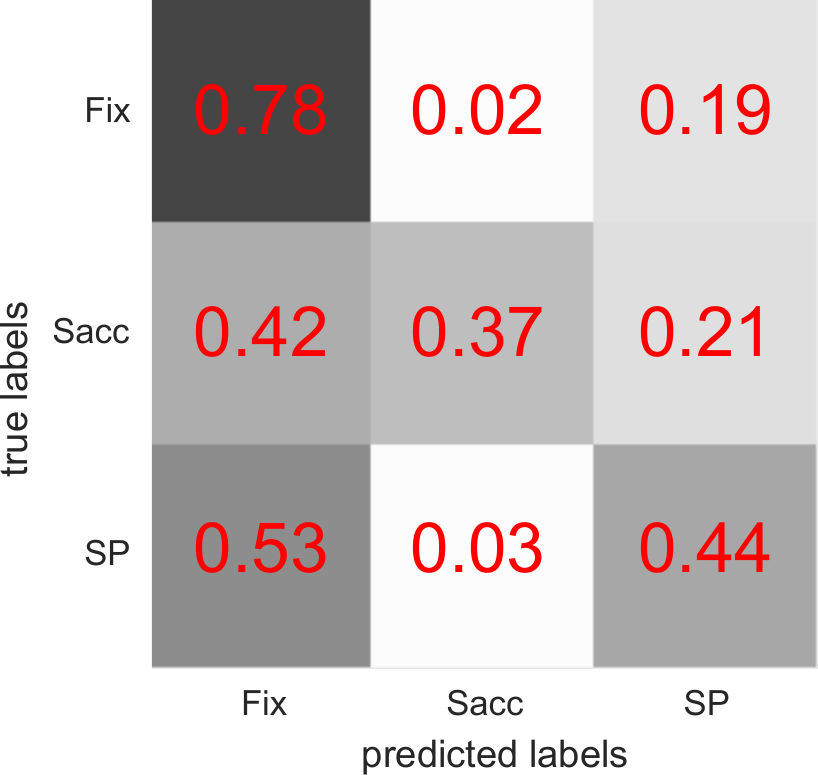}
\caption{Frame-wise confusion matrix in the multi-class setting. The numbers are normalised to sum up to one in each row, because the number of frames per class is imbalanced.}
\label{fig:confusion}
\end{figure}

Figure \ref{fig:confusion} shows the confusion matrix for the sample-by-sample evaluation. 
Fixations are labelled as fixations in 78\% of the cases, while for both pursuits and saccades there is a tendency to be labelled as fixations. They are correctly identified in 37\% and 44\% of frames respectively.
In turn, if a saccade is detected by the \cnn{}, it is almost certainly a true positive.
Table \ref{table:multi-class} summarises the performance of our classifier in terms of accuracy, precision, recall, and F1 score.

\def\arraystretch{1.3}%
\begin{table}[h]
\centering
\begin{tabular}{ccccc}
 & \textbf{Accuracy} & \textbf{Precision} & \textbf{Recall} & \textbf{F1 Score}\\
\toprule
Fixations & 0.65 & 0.63 & 0.78 & 0.70\\
Saccades & 0.87 & 0.77 & 0.37 & 0.50\\
Pursuits & 0.69 & 0.49 & 0.44 & 0.47\\
\hline
average & 0.74 & 0.63 & 0.53 & 0.55
\end{tabular}
\caption{Frame-wise performance in the multi-class setting.}
\label{table:multi-class}
\end{table}

Evaluating performance on event basis is difficult because it is unclear how to map the sequence of events predicted by our algorithm to the ground truth events. 
In particular, the \cnn{} tends to predict more events than there actually are, which makes it even harder to align.
One important question one can ask however, is how often the majority of predictions during one ground truth event belong to the correct class.
This problem could be rephrased as follows: if we had a segmentation given and did a simple majority voting in each segment, would we conclude the correct class label?

\def\arraystretch{1.3}%
\begin{table}
\centering
\begin{tabular}{c|ccc}
& \multicolumn{3}{c}{\textbf{Predictions}}\\
\textbf{Ground Truth}& Fixations & Saccades & Pursuits\\
\toprule
Fixations & 0.76 & 0 & 0.21 \\
Saccades  & 0.36 & 0.42 & 0.16 \\
Pursuits  & 0.41 & 0.01 & 0.54 \\
\end{tabular}
\begin{tabular}{ccc}

\end{tabular}
\caption{Event-based Evaluation: the numbers indicate the percentage of ground truth events for which the majorities of samples was labelled as a certain class. Where a line does not sum up to 1, this means that for some events, no label was predicted for more than half the frames.}
\label{table:eventsII}
\end{table}

Results for both of these evaluations can be found in Table \ref{table:eventsII}. 
For all three classes, we would predict the correct label more often than a false one.
Specifically, 76\% of fixations, 42\% of saccades and 54\% of pursuits would be identified correctly.
Comparing these results with the confusion matrix in Figure \ref{fig:confusion} illustrates how much a pre-segmentation simplifies the problem.

\subsection*{Prediction Confidence}

\begin{figure}
\centering
\includegraphics[width=.6\linewidth]{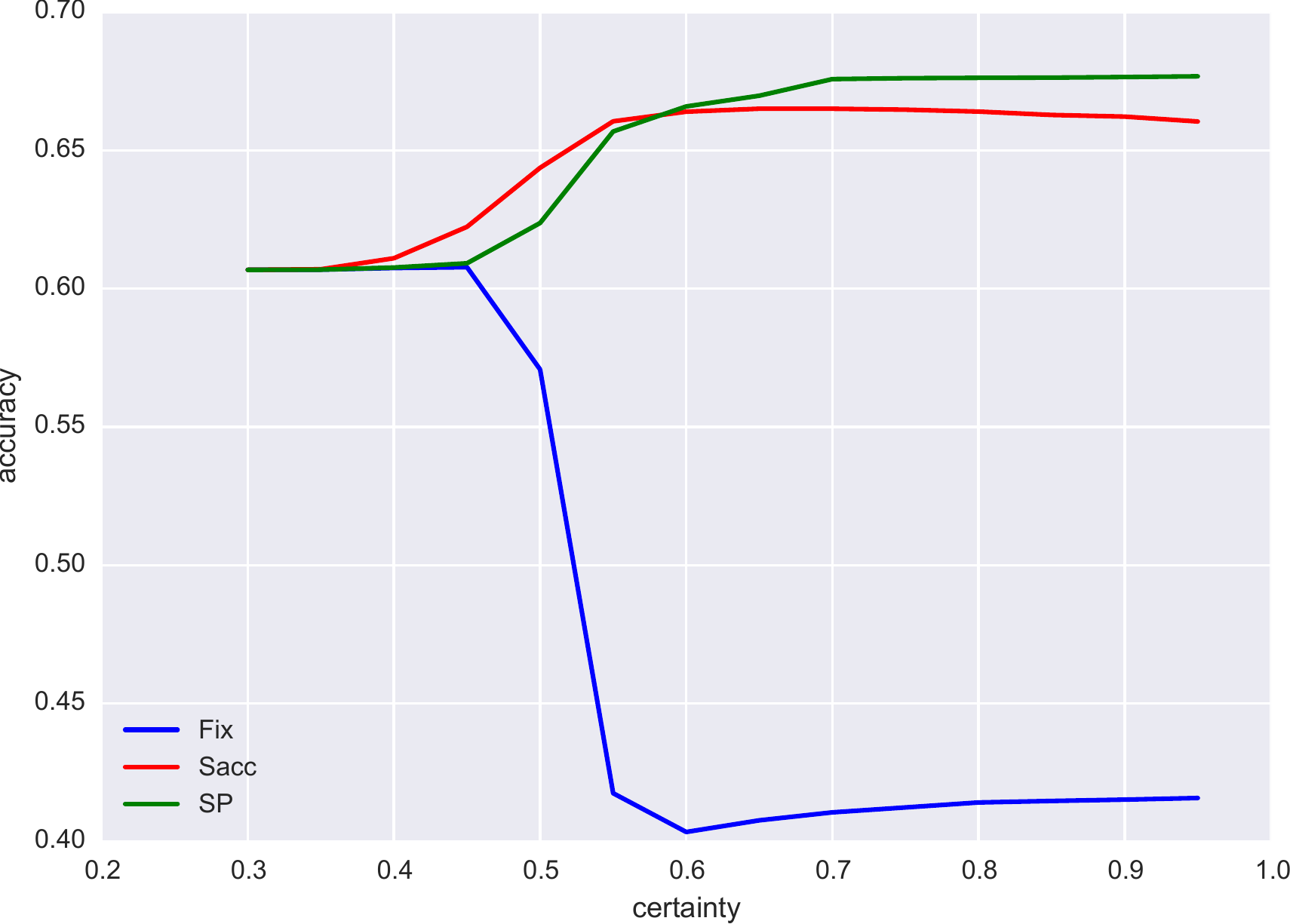}
\caption{Conditional probability of a prediction being correct given its certainty value.}
\label{fig:certainty}
\end{figure}

Given that there seems to be a general bias in the predictions towards fixations, the question arises how this relates to the probability values the \cnn{} computes for each prediction.
More precisely it would be a desirable property of the model if these probabilities corresponded to confidence, i.e. if a high probability would correlate with a high likelihood of the label for this instance being right.
Figure~\ref{fig:certainty} plots the accuracy of all predictions with a certain minimum probability associated. 
Each point in the graph corresponds to the accuracy (y axis) of predictions for a certain class (indicated by colours) given that the associated probability was at least the value indicated on the x axis.
As the red and green curves show, a label predicting a saccade or pursuit becomes more reliable with increasing probability, whereas only few movements classified as fixations actually correspond to fixations. 
This shows that the probabilities computed for a label can only be used as confidence values in the case of saccade and pursuit labels which raises new research questions such as how to use this information to improve detection results.

%% file: conclusion.tex
\section*{Discussion}

Our evaluation showed that the proposed \cnn{} outperforms existing approaches in binary eye movement classification and performs well on the significantly more challenging but practically clearly more relevant multi-class eye movement detection problem.
A comparison between sample-wise and event-based evaluation demonstrated how different these two metrics can look for one algorithm.
Sample-by-sample evaluation is inherently hard but we argue that it is nevertheless the correct evaluation setting for eye movement detection, in particular for sequence-to-sequence approaches.
While our \cnn{} based method outperforms existing methods our evaluations also show that robust and accurate eye movement detection is an inherently hard problem that poses interesting challenges for future research.
We compared our method with several state-of-the-art eye movement detection algorithms using a novel ground-truth annotated dataset of scripted and free-viewing eye movements.
Analysing scripted yet still natural eye movements proved to be a good approach to evaluate performance of eye movement detection algorithms in a principled way.

The proposed algorithm performs end-to-end detection of eye movements but still requires input gaze data to be transformed into the frequency domain first. 
While this transformation is computationally light-weight, it would be conceptually appealing to eliminate this step as well.
It will also be interesting to explore alternatives to the proposed FFT transformation, such as wavelets that are used in EOG data processing ~\cite{bulling11_pami}, could improve over the results presented here.

In computer vision, much of the success of convolutional neural networks only started after large amounts of data became available and the networks could become more complex without overfitting (cf. ~\cite{dean2012large}).
In comparison to large databases available in computer vision, such as ~\cite{imagenet} with more than 14 million images, the dataset introduced here is still small. 
Hence it will be interesting to study how \cnns{} will evolve with more labelled data becoming available for training.
Generating such large datasets will be a significant effort since gaze data for non-scripted stimuli needs to be manually annotated.
In addition, it will be interesting to investigate if unlabelled data can be used -- again, similar in spirit to existing computer vision works~\cite{vidya2014sparse}.

Although methods for inspecting the internal structures learned by convolutional neural networks continuously improve, the networks are currently still known to be difficult to interpret~\cite{zeiler2014visualizing}.
One thing we do know about convolutional neural networks is that they do not have an internal state, i.e.\ they do not remember the last data samples they have seen.
However, gaze data is inherently temporal on several scales and has repetitive characteristics that could be exploited for eye movement detection.
For example, one gaze sample can only deviate a certain amount from the previous due to physiological restrictions ~\cite{robinson1965mechanics} and many of the factors influencing gaze characteristics only change slowly over time, e.g. fatigue.
Thus, another promising direction for future research is to investigate alternative neural network architectures with memory~\cite{hochreiter1997long}.

\section*{Conclusion}

In this work we have proposed a convolutional neural network to detect fixations, saccades and smooth pursuits from a stream of gaze data.
To the best of our knowledge, this is the first approach that can detect all three classes simultaneously from a continuous sequence of gaze data as returned by current eye trackers. 
Convolutional neural networks are capable of learning suitable representations for the given data on their own, without manual feature crafting. 
Thus, this approach learns the entire process of detection from the input to the label with associated probability values end to end. 
In particular, no explicit segmentation is required nor is an implicit segmentation step included in the algorithm.

Moreover, we have introduced a new fully annotated dataset containing gaze data from 15 people for several scripted and free-viewing tasks as well as reading. In our evaluation we have shown that despite the superiority of our approach over baseline methods, eye movement detection remains a hard problem requiring further investigations.